\documentclass{article}

\PassOptionsToPackage{round}{natbib}

\usepackage[preprint]{neurips_2022}

\usepackage{hyperref}
\usepackage{algorithm}
\usepackage{algorithmic}

\usepackage[utf8]{inputenc} 
\usepackage[T1]{fontenc}    
\usepackage{url}            
\usepackage{booktabs}       
\usepackage{amsfonts}       
\usepackage{nicefrac}       
\usepackage{microtype}      
\usepackage{xcolor}         

\usepackage{amsmath}
\usepackage{amssymb}
\usepackage{mathtools}
\usepackage{amsthm}

\usepackage[capitalize,noabbrev]{cleveref}

\theoremstyle{plain}

\theoremstyle{definition}

\theoremstyle{remark}

\usepackage{url}

\usepackage[load=prefixed,load=abbr,separate-uncertainty=true]{siunitx} 

\newcommand{\sub}[1]{_{\mathrm{#1}}}

\newcommand{\pmstack}[2]{\substack{+#1 \\ -#2}}

\newcommand{\ie}{i.e.\@}
\newcommand{\eg}{e.g.\@}
\newcommand{\bigO}[1]{\mathcal{O}(#1)}
\newcommand{\smref}[1]{SM #1}

\usepackage{subfigure}
\usepackage{lineno}
\usepackage[T1]{fontenc}
\usepackage{graphicx}
\usepackage{wrapfig}
\usepackage{physics}
\usepackage{siunitx}

\usepackage{booktabs}
\newcommand{\tabscale}{0.9}

\usepackage{xcolor}

\newcommand{\sota}{SOTA}
\newcommand{\maxcut}{Max-Cut}
\newcommand{\tabref}[1]{table~\ref{#1}}

\newcommand{\R}{\mathbb{R}}
\newcommand{\Rab}[2]{\mathbb{R}^{#1{\times}#2}}
\DeclareMathOperator{\layernorm}{LN}
\DeclareMathOperator*{\argmax}{arg\,max}
\newcommand{\softmax}{\mathrm{softmax}}

\title{Learning to Solve Combinatorial Graph Partitioning Problems via Efficient Exploration}

\author{%
  Thomas D.~Barrett \\
  InstaDeep\\
  \And
  Christopher W.F.~Parsonson\thanks{Work completed during internship at InstaDeep.} \\
  University College London, UK\\
  \And
   Alexandre Laterre \\
  InstaDeep\\
}

\begin{document}

\maketitle

\begin{abstract}
From logistics to the natural sciences, combinatorial optimisation on graphs underpins numerous real-world applications.
Reinforcement learning (RL) has shown particular promise in this setting as it can adapt to specific problem structures and does not require pre-solved instances for these, often NP-hard, problems.
However, state-of-the-art (\sota{}) approaches typically suffer from severe scalability issues, primarily due to their reliance on expensive graph neural networks (GNNs) at each decision step.
We introduce ECORD; a novel RL algorithm that alleviates this expense by restricting the GNN to a single pre-processing step, before entering a fast-acting exploratory phase directed by a recurrent unit.
Experimentally, ECORD achieves a new \sota{} for RL algorithms on the Maximum Cut problem, whilst also providing orders of magnitude improvement in speed and scalability.  Compared to the nearest competitor, ECORD reduces the optimality gap by up to \SI{73}{\percent} on \num{500} vertex graphs with a decreased wall-clock time.  Moreover, ECORD retains strong performance when generalising to larger graphs with up to \num{10000} vertices.
\end{abstract}

\section{Introduction}

Combinatorial optimisation (CO) problems seek an ordering, labelling, or subset of discrete elements that maximises some objective function.  Despite this seemingly abstract mathematical formulation, CO is at the heart of many practical applications, from logistics \citep{Yanling_2010} to protein folding \citep{perdomo12} and fundamental science \citep{barahona82}.  However, with such tasks often being NP-hard, solving CO problems becomes increasingly challenging for all but the simplest of systems.
This combination of conceptual simplicity, computational complexity, and practical importance has made CO a canonical challenge, and motivated significant efforts into developing approximate and heuristic algorithms for these tasks.
Whilst approximate methods can offer guarantees on the solution quality, in practice they frequently lack sufficiently strong bounds and have limited scalability \citep{williamson11}.  By contrast, heuristics offer no such guarantees but can prove highly efficient~\citep{Halim2019CombinatorialOC}.

As a result, recent years have seen a surge in the application of automated learning methods that parameterise CO heuristics with deep neural networks \citep{bengio2020machine}.
In particular, reinforcement learning (RL) has become a popular paradigm, as it can facilitate the discovery of novel heuristics without the need for labelled data.
Moreover, many CO problems are naturally formulated as Markov decision processes (MDPs) on graphs, where vertices correspond to discrete variables and edges denote their interaction or dependence.
Accordingly, graph neural networks (GNNs) have become the de-facto function approximator of choice as they reflect the underlying structure of the problem whilst seamlessly handling variable problem sizes and irregular topological structures. 

However, despite the demonstrated success of RL-GNN approaches, scalability remains an outstanding challenge.  Running a GNN for every decision results in impractical computational overheads for anything beyond small- to medium-sized problems.
This is exacerbated by the fact that directly predicting the solution to an NP-hard problem is typically unrealistic.
Indeed, recent approaches often utilise stochastic exploration or structured search to generate multiple candidate solutions \citep{chen2019learning, joshi2019efficient, gupta2020hybrid, barrett20, bresson2021transformer}, which further increases the computational burden.
In this work, we look to address these challenges by developing a combinatorial solver that leverages the performance benefits of GNN function approximators and search-based approaches, whilst remaining computationally scalable.

Our intuition is that an agent trained to explore the solution space should be able reason over both the geometric structure of the problem (how the discrete elements interact) and the temporal structure of the search (what solutions have been previously explored).
In contrast to prior works that focus on expensive structural reasoning over the same problem instance at each step~\citep{Dai_2017, barrett20}, we introduce a new algorithm, ECORD (Exploratory Combinatorial Optimisation with Recurrent Decoding),
that combines a single GNN preprocessing step with fast action-decoding -- replacing further geometric inference with simple per-vertex observations and a learnt representation of the ongoing optimisation trajectory.
The result is a performant CO solver with a theoretical and demonstrated speed-up over expensive GNN action-decoding -- indeed, the action-selection time of ECORD is independent of the graph topology and, in practice, near constant regardless of graph size.

Experimentally, we consider the Maximum Cut (\maxcut) graph partitioning problem, chosen because of its generality (11 of the 21 NP-complete problems of \citet{karp72} can be reduced to \maxcut{}, including graph coloring, clique cover, knapsack and Steiner Tree) and the fact that it is a challenging problem for scalable CO as every vertex must be correctly labelled (rather than simply a subset).
Indeed, there has been significant commercial and research efforts into \maxcut{} solvers, from bespoke hardware based on classical~\citep{goto19} and quantum annealing~\citep{yamamoto17, djidjev18} to hand-crafted~\citep{goemans95, benlic13} or learnt~\citep{barrett20} algorithms.
These efforts stand as testament to the combination of intractability and broad applicability that motivates our choice of problem class.

ECORD is found to equal or surpass the performance of expensive \sota{} RL-GNN methods on graphs with up to \num{500} vertices (where all methods are computationally feasible), even when compared on number of actions instead of wall-clock time.
Moreover, ECORD provides orders of magnitude improvements in speed and scalability, with strong performance and a nearly $300{\times}$ increase in throughput when compared to conventional RL-GNN methods, on graphs with up to \num{10000} vertices.

\section{Related work}

Whilst tackling CO problems with neural networks \citep{hopfield85} and RL \citep{zhang_1995} is not a new idea, recent years has seen a resurgence of interest in developing CO solvers with ML, with multiple reviews providing detailed taxonomies \citep{bengio2020machine,mazyavkina2020reinforcement, vesselinova_2020}.

\paragraph{Learning to solve non-Euclidean CO}
This resurgence began by considering Euclidean approaches that did not reflect the underlying graph structure of the problems.
In this context, \citet{bello16} used RL to train pointer networks (PNs), which treat the discrete variables of the CO problem as a sequence of inputs to a recurrent unit~\citep{vinyals15} to solve TSP.
\citet{Gu_2020} further scaled this approach to instances of up to \num{300} vertices, using a hybrid supervised-reinforcement learning framework that combined PNs and A3C \citep{mnih2016asynchronous}. However, these Euclidean approaches fail to capture the topological structures and intricate relationships contained within graphs, and typically require a large number of training instances in order to generalise.
This was addressed by \citet{Dai_2017}, who trained a Structure-to-Vector GNN with DQN to solve TSP and \maxcut{}. The resulting algorithm, S2V-DQN, generalised to graphs with different size and structure to the training set and achieved excellent performance across a range of problems without the need for manual algorithmic design, demonstrating the value in exploiting underlying graph structure.

\paragraph{Advances in optimality}
Various works since \citet{Dai_2017} have sought to harness GNN embeddings to improve solution quality. \citet{abe19} combined a GNN with a Monte Carlo tree search (MCTS) approach to learn a high-quality constructive heuristic. Ultimately, they demonstrated weaker performance but a greater ability to generalise to unseens graph types than S2V-DQN on \maxcut{}.
\citet{li18} combined a graph convolution network (GCN) with guided tree search to synthesise a diverse set of solutions and thereby more fully explore the space of possible solutions. However, they used supervised learning and so required labelled data which limits scalability, and they did not consider \maxcut{}.
\citet{barrett20} proposed ECO-DQN, the \sota{} RL algorithm for \maxcut{}, which reframed the role of the RL agent as looking to improve on any given solution, rather than directly predict the optimal solution.  The key insight is that exploration \textit{at test time} is beneficial, since sub-optimal decisions, which are to be expected for NP-hard problems, can be revisited. However, ECO-DQN utilises a costly GNN at each decision step -- which, coupled with exploratory trajectories during inference, results in scalability even worse than that of S2V-DQN -- and represents the previous search trajectory with only a few hand-crafted features.
ECORD remedies this by using an initial GNN embedding followed by a recurrent unit to balance the richness provided by graph networks with fast-action~selection conditioned on a learnt representation of the trajectory.

\paragraph{Advances in scalability}
\citet{manchanda2020learning} furthered the work of \citet{Dai_2017} by first training an embedding GCN in a supervised manner, and then training a Q-network with RL to predict per-vertex action values. By using the GCN embeddings to prune nodes unlikely to be in the solution set, their method provided significantly more scalability than S2V-DQN on the Maximum Vertex Cover problem. However, it is not applicable to problems where nodes cannot be pruned, which includes some of the most fundamental CO problems such as \maxcut{}.
\citet{drori2020learning} proposed a general framework that uses a graph attention network to create a dense embedding of the input graph followed by an attention mechanism for action selection. They achieved impressive scalability, reaching instance sizes of \num{1000} vertices on the minimum spanning tree problem and TSP. However, unlike ECORD,  the \citet{drori2020learning} framework restricts the decoding stage to condition only on the previously selected action, and considers only node ordering problems in a non-exploratory setting.
\section{Methods}

\subsection{Background}
\label{sec:background}

\paragraph{\maxcut{} problem}
The \maxcut{} problem seeks a binary labelling of all vertices in a graph, such that the cumulative weight of edges connecting nodes of opposite labels is maximised.  Concretely, for a graph, $G(V,E)$, with vertices $V$ and edges $E$, we seek a subset of vertices, $S {\subset} V$ that maximises the `cut value', $C(S,G) {=} \sum_{i {\in} S, j {\in} V {\setminus} S} w(e_{ij})$, where $w(e_{ij})$ is the weight of an edge $e_{ij} {\in} E$.

\paragraph{Q-learning} 
We formulate the optimisation as a Markov Decision Process (MDP) $\{\mathcal{S}, \mathcal{A}, \mathcal{T}, \mathcal{R}, \gamma \}$, where $\mathcal{S}$ and $\mathcal{A}$ are the state and action spaces, respectively, $\mathcal{T}: \mathcal{S}{\times}\mathcal{A}{\times}\mathcal{S} \rightarrow [0,1]$ is the transition function,  $\mathcal{R}:  \mathcal{S} \rightarrow \mathbb{R}$ the reward function and $\gamma \in [0,1]$ the discount factor.
Q-learning methods learn the Q-value function mapping state-action pairs, $( s {\in} \mathcal{S}, a {\in} A )$, to the expected discounted sum of rewards when following a policy $\pi{:} \mathcal{S} {\rightarrow} \mathcal{A}$,
$Q^{\pi}(s, a) = \mathbb{E}_{\pi} [ \sum_{t^\prime=t+1}^{\infty} \gamma^{t^\prime-1} r(s^{(t^\prime)}) \vert s^{(t)} {=} s, a^{(t)} {=} a ]$.
DQN~\citep{mnih15} optimises $Q_\theta$ by minimising the error between the network predictions and a bootstrapped estimate of the Q-value. After training, an approximation of an optimal policy is obtained by acting greedily with respect to the learnt value function, $\pi_{\theta}(s) = \argmax_{a^{\prime}} Q_\theta(s, a^{\prime})$.

\paragraph{Munchausen DQN}
Munchausen DQN (M-DQN)~\citep{vieillard20} makes two fundamental adjustments to conventional DQN.  Firstly, the Q-values are considered to define a stochastic policy with action probabilities given by $\pi_{\theta}(\cdot \vert s) = \softmax(\tfrac{Q_\theta(s,\cdot)}{\tau})$, where $\tau$ is a temperature parameter.  Secondly, M-DQN adds the log-probability of the selected action to the reward at each step, with scaling parameter $\alpha$.  All together, the regression target for the Q-function is modified to
\begin{equation}
		Q\sub{mdqn}
	{=} r (s^{(t+1)})
	{+} \alpha\tau \ln\pi_\theta(a^{(t)} \vert s^{(t)})
	{+} \gamma \mathbb{E}_{a^{\prime} \sim \pi( \cdot | s^{(t+1)})} [ Q_\theta(s^{(t+1)}, a^{\prime})
	{-} \tau \ln\pi_\theta(a^{\prime} \vert s^{(t+1)})  ].
\end{equation}

\subsection{ECORD}
\label{sec:ecord}

\paragraph{MDP formulation}
The state, $s^{(t)} \equiv (G(V,E), S) \in \mathcal{S}$, at a given time-step corresponds to the graph and a binary labelling of all vertices.
An action selects a single vertex and `flips' its label to the opposite state, adding or removing it from the solution subset.
Rewards correspond to the (normalised and clipped) increase in best cut-value, $r(s^{(t)}) = \max( ( C(s^{(t)}) - C(s^{\ast}) ) / \abs*{V}, 0 )$, where $C(s^{\ast})$ is the the best solution found in the episode thus far.  This choice of reward structure motivates the agent to maximise the cut-value at any step within the episode, without discouraging exploration.

Concretely, we represent the state, $s^{(t)}$, with per-vertex observations, $o_i^{(t)} {\in} \mathbb{R}^{3}$ for $i {\in} V$, and a global observation of the current state, $o\sub{G}^{(t)} {\in} \mathbb{R}^{2}$.
For each vertex, we provide: (i)~the current label, (ii)~the immediate change in cut-value if the vertex is flipped (which we refer to as a `peek') and (iii)~the number of steps since the node was last flipped.  Our global observations are the (normalised) difference in cut-value between the current and best observed state, and the maximum peek of any vertex in the current state.
Importantly, these features are readily available without introducing any significant computational burden. Even the peeks -- one-step look-aheads for each action -- can be calculated for all vertices once at the beginning of each episode, after which they can be efficiently updated  at each step for only the flipped vertex and it's immediate neighbours (details can be found in \smref{Section C}).  Besides being a cheap operation, this approach is also the most efficient way to track the cut-value over the course of multiple vertex flips, and thus is a necessary calculation to evaluate each environmental state regardless of whether the peeks form part of the observation.

\paragraph{Architecture}
\begin{figure*}[!t]
	\centering
	\includegraphics[width=0.8\textwidth]{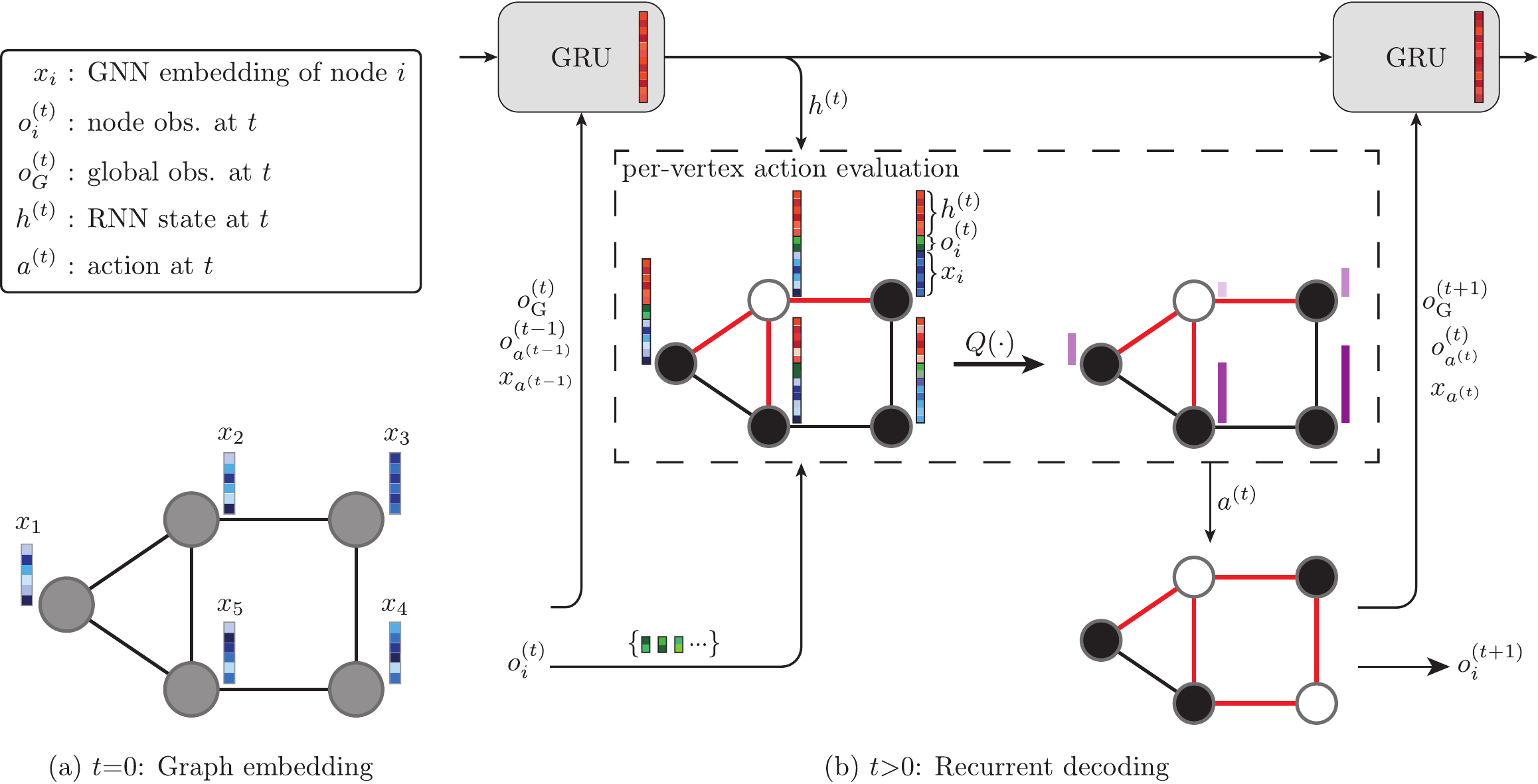}
    \caption{The architecture of ECORD. (a)~A GNN generates embeddings for each vertex at $t=0$. (b)~At each timestep, a recurrent unit takes as input the GNN embedding, the node-level observation from the previous time step, and the global observation from the current time step, and updates its hidden state. To select an action (node to flip), the hidden state is concatenated with the node-level observation at the current time step and passed to a Q-network.
    }
    \label{fig:architecture}
\end{figure*}
The architectural contribution of ECORD is demonstrating that performant exploration is possible when combining a single GNN pre-processing step with recurrent decoding that maintains a representation of the ongoing optimisation trajectory.  Concretely, ECORD splits an optimisation into two stages (see \cref{fig:architecture} for an overview of the architecture).
(\romannumeral 1) A GNN prepares per-vertex embeddings conditioned on only the geometric structure (\ie{} weighted adjacency matrix) of the graph.
(\romannumeral 2) Starting from a random labelling, ECORD sequentially flips vertices between subsets, with each action conditioned on the static GNN embeddings, the previous trajectory (through the hidden state of an RNN) and simple observations of the current state (described above).
Full architectural details can be found in \smref{Section B}, with this section providing a higher level summary.

The per-vertex embeddings are generated with a gated graph convolution network~\citep{li15}.  The final embeddings after $L$ rounds of message-passing are linearly projected to per-vertex embeddings, $x_i = W\sub{p} x_i^{(L)}$ where $i$ denotes the vertex, that remain fixed for the rest of the episode. 
To select an action at time $t$, we first combine the GNN outputs with the observation to obtain per-vertex embeddings of the form $v_i^{(t)} = [ x_i, W\sub{o} o_i^{(t)} ]$, where square brackets denote concatenation.
This local information is then used in conjunction with the RNN hidden state, $h^{(t)}$, to predict the Q-value for flipping vertex $i$.  The Q-network itself uses a duelling architecture~\citep{wang16}, where separate MLPs estimate the value of the state, $V(\cdot)$, and advantage for each action, $A(\cdot)$,
\begin{equation}
	Q(v_i^{(t)}, h^{(t)}) = \text{MLP}\sub{V}(h^{(t)}) {+} \text{MLP}\sub{A}([v_i^{(t)}, W\sub{h} h^{(t)}]).
\label{eq:qVals}
\end{equation}
$h^{(t)}$ is shared across each vertex when calculating the advantage function as our reward function depends on the current best score within the trajectory's history, a global feature contained in the state-level embedding.  Before concatenating them together, we project the high (1024) dimensional hidden state to match the lower (32)  dimensional per-vertex embeddings.
Finally, the RNN is updated using the embedding of the selected action, $v_{\ast}^{(t)}$, and the next global observation, $o\sub{G}^{(t+1)}$,
\begin{equation}
	h^{(t+1)} = \text{GRU}(h^{(t)}, \text{MLP}\sub{M}([v_{\ast}^{(t)}, o\sub{G}^{(t+1)}])).
\end{equation}

\paragraph{Training}
ECORD is trained using M-DQN where (recall \cref{sec:background}): (i) the entropy of the policy is jointly optimised with the returns, in the spirit of maximum entropy RL~\citep{haarnoja18}, and
(ii) the agent is rewarded for being more confident about the correct action.
One could observe that as ECORD's action space is large (of size $\abs*{V}$) the structured exploration of M-DQN's stochastic policy may allow for more meaningful trajectories than the standard (epsilon-)greedy approaches used in DQN.
Indeed, as we will see, leveraging the stochasticity of ECORD's policy is important for maximising performance at inference time.
As the Q-values are conditioned upon the internal state of an RNN, see eq.~(\ref{eq:qVals}), we train using truncated backpropagation through time (BPTT).  Concretely, when calculating the Q-values at time $t$ during training, we reset the environment and the internal state of the RNN to their state at time $t - k\sub{BPTT}$ and replay the trajectory to time $t$.  In doing so, we only have to backpropagate through the previous $k\sub{BPTT}$ time steps when minimising the loss.
\section{Experiments}

Our experiments first investigate and analyse key components of ECORD.  Concretely, we compare the quality of decisions made by our efficient architecture to those made by expensive per-step RL-GNN approaches (\cref{sec:action_quality}), demonstrate ECORD's theoretical and practical improvements in run-time and scalability (\cref{sec:complexity}), and highlight the importance of learning a stochastic policy (\cref{sec:gset}).
Finally, we combine everything into a single performant CO solver, ECORD, and investigate its performance and ablations on large problem instances (\cref{sec:gset}).
Supporting code and all datasets considered in this work are either available or linked at \url{github.com/tomdbar/ecord}.

\paragraph{Baselines}
We compare ECORD to the previous two \sota{} RL-GNN algorithms for \maxcut{}, S2V-DQN~\citep{khalil17} and ECO-DQN~\citep{barrett20}.  Our implementation of ECORD contains several speed improvements in comparison to the publicly available implementation of ECO-DQN 
(\eg{}~parallelised optimisation trajectories, compiled calculations and sparse matrix operations). To ensure the fairest possible comparison,
when directly comparing performance independent of speed (Section~\ref{sec:action_quality}), we use the public implementation, however when comparing scalability (Section~\ref{sec:complexity} and \ref{sec:gset}) we use a re-implementation of ECO-DQN within the same codebase as ECORD.

Additionally, a simple heuristic, \maxcut{} Approx (MCA), that greedily selects actions that maximise the immediate increase in cut value is also considered.  Besides from providing surprisingly strong performance, the choice of MCA baselines is motivated by the observation that the one-step look-ahead `peek' features make learning a greedy policy straightforward.  Moreover, whilst MCA terminates once a locally optimal solution is found (\ie{} one from which no single action can further increase the cut value), a network learning to only approximate a greedy policy may fortuitously escape these solutions and ultimately obtain better results.  To address this, we introduce a simple extension (and significant improvement) of MCA called MCA-soft, which uses a soft-greedy policy with the temperature tuned to maximise performance on the target dataset (see \smref{Section D}).

\begin{table*}[!t]
\caption{
 Performance of agents trained on ER40 with $2\abs{V}$ actions per episode (best in \textbf{bold}). Error bars denote \SI{68}{\percent} confidence intervals. RL baseline scores are taken directly from~\citet{barrett20}.  Results are averaged across 5 seeds for the RL baselines and 3 seeds for ECORD-det.
}
\label{tab:sota_comp}
\vskip 0.15in
\centering
\scalebox{\tabscale}{%
\begin{tabular}{c cc cc c}

\toprule

\multicolumn{1}{c}{} &
\multicolumn{2}{c}{Heuristics} &
\multicolumn{2}{c}{RL baselines} &
\multicolumn{1}{c}{}
\\

\cmidrule[0.75pt](lr){2-3}
\cmidrule[0.75pt](lr){4-5}

& MCA & MCA-soft & S2V-DQN & ECO-DQN & ECORD-det \\

\cmidrule[0.75pt](lr){1-6}

ER40 &
 $0.997\pmstack{0.001}{0.010}$ & 
 $\mathbf{ 1.000 }\pmstack{0.000}{0.000}$ & 
 $0.980\pmstack{0.014}{0.023}$ &
 $\mathbf{ 1.000 }\pmstack{0.000}{0.000}$ & 
 $\mathbf{ 1.000 }\pmstack{0.000}{0.000}$ \\
ER60 &
 $0.994\pmstack{0.003}{0.013}$ & 
 $0.999\pmstack{0.001}{0.004}$ & 
 $0.973\pmstack{0.021}{0.024}$ &
 $\mathbf{ 1.000 }\pmstack{0.000}{0.000}$ &
 $\mathbf{ 1.000 }\pmstack{0.000}{0.000}$ \\
ER100 &
 $0.977\pmstack{0.017}{0.017}$ &
 $0.993\pmstack{0.007}{0.008}$ &
 $0.961\pmstack{0.029}{0.028}$ &
 $\mathbf{ 1.000 }\pmstack{0.000}{0.001}$ &
 $\mathbf{ 1.000 }\pmstack{0.000}{0.001}$ \\
ER200 &
 $0.959\pmstack{0.014}{0.014}$ &
 $0.973\pmstack{0.012}{0.012}$ &
 $0.951\pmstack{0.020}{0.020}$ &
 $0.999\pmstack{0.001}{0.002}$ &
 $\mathbf{ 1.000 }\pmstack{0.000}{0.001}$\\
ER500 &
 $0.941\pmstack{0.012}{0.012}$ &
 $0.958\pmstack{0.009}{0.009}$ & 
 $0.921\pmstack{0.019}{0.019}$ &
 $0.985\pmstack{0.006}{0.006}$ &
 $\mathbf{ 0.996 }\pmstack{0.004}{0.004}$ \\
 
\cmidrule[0.5pt](lr){1-6}
 
BA40 &
 $0.999\pmstack{0.000}{0.006}$ & 
 $\mathbf{ 1.000 }\pmstack{0.000}{0.000}$ & 
 $0.967\pmstack{0.023}{0.040}$ &
 $\mathbf{ 1.000 }\pmstack{0.000}{0.000}$ & 
 $\mathbf{ 1.000 }\pmstack{0.000}{0.000}$ \\
BA60 &
 $0.989\pmstack{0.007}{0.018}$ & 
 $0.997\pmstack{0.003}{0.008}$ & 
 $0.968\pmstack{0.022}{0.036}$ &
 $\mathbf{ 1.000 }\pmstack{0.000}{0.000}$ &
 $\mathbf{ 1.000 }\pmstack{0.000}{0.000}$ \\
BA100 &
 $0.965\pmstack{0.020}{0.020}$ &
 $0.984\pmstack{0.012}{0.013}$ &
 $0.940\pmstack{0.032}{0.033}$ &
 $\mathbf{ 1.000 }\pmstack{0.000}{0.000}$ &
 $\mathbf{ 1.000 }\pmstack{0.000}{0.000}$ \\
BA200 &
 $0.911\pmstack{0.033}{0.037}$ &
 $0.929\pmstack{0.034}{0.034}$ &
 $0.865\pmstack{0.058}{0.061}$ &
 $0.978\pmstack{0.014}{0.033}$ &
 $\mathbf{ 0.983 }\pmstack{0.017}{0.033}$ \\
BA500 &
 $0.889\pmstack{0.015}{0.015}$ &
 $0.899\pmstack{0.014}{0.014}$ & 
 $0.744\pmstack{0.052}{0.052}$ &
 $\mathbf{ 0.967 }\pmstack{0.014}{0.015}$ &
 $0.963\pmstack{0.012}{0.012}$ \\
  
\bottomrule

\end{tabular}
} 
\vskip -0.1in
\end{table*}

\paragraph{Datasets}
We consider graph datasets for which the optimal (or best known) solutions are publicly available.  The dataset published by \citet{barrett20} consists of Erd\H{o}s-R\'{e}nyi~\citep{erdos60} and Barabasi-Albert ~\citep{albert02} graphs (ER and BA, respectively) with edge weights $w(e_{ij})\in\{0,{\pm}1\}$ and up to 500 vertices.  Each graph type and size is represented by 150 random instances, with 50 used for model selection and results reported on the remaining 100 at test time.  We refer to these distributions as ER40/BA40 to ER500/BA500.

To test on larger graphs we use the GSet~\citep{benlic13}, a well-studied dataset containing multiple distributions, from which we focus on random (ER) graphs with binary edge weights and 800 to \num{10000} vertices.  To aid model selection and parameter tuning, we generate 10 additional graphs for each distribution in the GSet. We refer to these validation sets as ER800 to ER10000.

\paragraph{Metrics}
Our analysis considers the raw performance, wall-clock time, and memory usage of algorithms.  Following prior work, we use the approximation ratio, given by $AR(s^{\ast}) = C(s^{\ast}) / C\sub{opt}$ where $C\sub{opt}$ is the best known cut value, as a metric of solution quality.
All experiments were performed on the same system with an Nvidia GeForce RTX 2080 Ti \SI{11}{\giga B} GPU and 80 processors (Intel(R) Xeon(R) Gold 6248 CPU @ \SI{2.50}{\GHz}).

\subsection{Action quality of ECORD}
\label{sec:action_quality}

ECORD restricts the GNN to only a single pre-processing step and so, in contrast to SOTA RL-GNN approaches, does not directly condition per-vertex decisions on the state of other vertices.  One could then ask whether the actions ECORD takes are less informed or, even, if they could be replicated with simple per-vertex policies that ignore the graph structure entirely.  Therefore, we first investigate the per-decision quality of ECORD's architecture, whilst ignoring any scalability benefits.

\paragraph{Methods}
ECORD is compared to both heuristic vertex-flipping, and \sota{} RL, baselines (specifically, ECO-DQN and S2V-DQN).
All RL algorithms are trained on $\abs*{V}{=}40$ ER graphs and evaluated on both ER and BA graphs with up to $\abs*{V}{=}500$.  S2V-DQN is deterministic and constructs the solution set one vertex at a time, therefore each optimisation trajectory consists of $\abs*{V}$ sequential actions.   MCA-soft, ECO-DQN and ECORD allow any vertex to be `flipped' at each step, and therefore can in principle run indefinitely on a target graph.
The baselines presented are chosen as they consider the same sequential node-flipping paradigm as ECORD, however their exists multiple algorithms for solving \maxcut{} that do not fit this paradigm -- notably simulated annealing (SA)~\citep{tiunov19,leleu19}, semidefinite programming (SDP)~\citep{goemans95} and mixed integer programming~\citep{cplex07}.  We provide additional results using SOTA or commercial algorithms spanning these paradigms in \smref{Section E}, where ECORD is found to outperform SDP and MIP and either beat or be competitive with \sota{} SA methodologies.

In practice, we use ECORD and the MCA heuristics with ECO-DQN's default settings; 50 optimisation trajectories per graph, each starting from a random node labelling, acting greedily with respect to the learnt Q-values, and terminating after $2\abs*{V}$ sequential actions.  We note that this disadvantages ECORD as
(\romannumeral 1)~despite learning a stochastic policy, ECORD acts deterministically at test time, and
(\romannumeral 2)~ECORD's significant speed and memory advantages over ECO-DQN are not accounted for.
To distinguish this deterministic implementation from a full-strength version of ECORD (which we benchmark on more challenging instances in \cref{sec:gset}), we refer to it as ECORD-det.

\paragraph{Results}
Despite these disadvantages, ECORD-det either outperforms or essentially matches ECO-DQN on all tests (see \tabref{tab:sota_comp}), whilst significantly improving over all other baselines and generalising to unseen graph topologies and sizes.  
Although, in contrast to the RL baselines, ECORD-det does not directly condition per-vertex decisions on the state of other vertices, it does have access to the optimisation trajectory via the RNN hidden state.  ECORD-det's strong performance then evidences our intuition that, in exploratory settings, the temporal structure (in addition to the geometric structure of the graph) is highly informative.
Ultimately, on these small instances where some baseline's are already near-optimal, we are not investigating the full performance of ECORD, but rather verifying that our proposed architecture remains capable of taking high quality decisions on par with more computationally expensive methods.
This experiment was repeated for a model trained instead on BA40 with the results (provided in \smref{Section E}) being qualitatively the same.

\subsection{Computational complexity and practical scaling}
\label{sec:complexity}

\begin{figure*}[!b]
     \centering
	\begin{subfigure}
         \centering
         \includegraphics[width=0.4\textwidth]{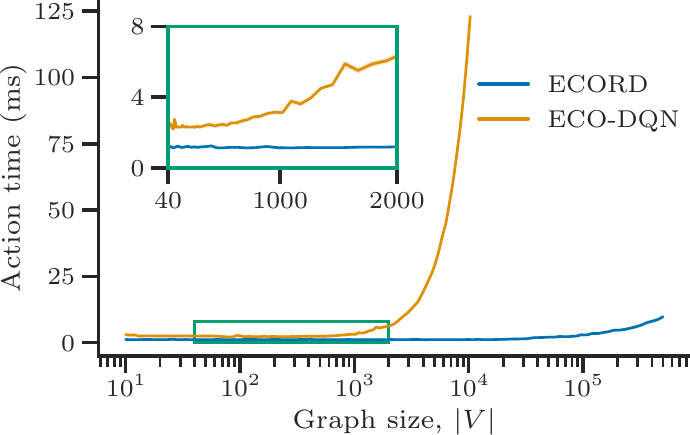}
         \label{fig:speed_comparison}
     \end{subfigure}
     \hspace{0.025\textwidth}
	\begin{subfigure}
	     \centering
         \includegraphics[width=0.4\textwidth]{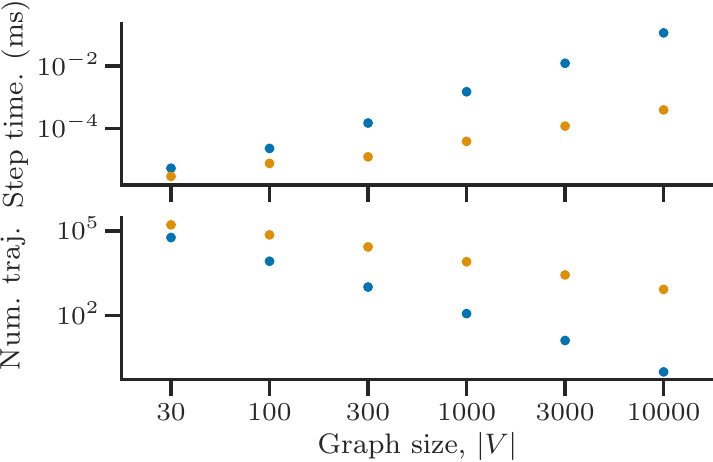}
         \label{fig:throughput_comparison}
     \end{subfigure}
     \caption{ECORD (blue) vs ECO-DQN (orange) scaling performance.  (left) Action selection time when running a single trajectory. (right) Step time (action selection and environment update time) when running the maximum number of tradjectories that fit on the GPU.  ECORD's relative throughput at $|V|=\{30,100,300,\num{1000},\num{3000},\num{10000}\}$ is $\{1.8, 3.0, 12.2, 39.3, 104.5, 298.9\}\times$ ECO-DQN's (beyond $|V|{\approx}\num{10000}$, ECO-DQN no longer fits on the GPU).}
\label{fig:performance_comparison}
\end{figure*}

Whilst ECORD has already been shown to match or surpass \sota{} RL baselines when ignoring computational cost, this is not a sufficient metric of algorithmic utility when a key aim of ECORD is to scale exploratory CO to large instances.  In this section, we consider the theoretical complexity and practical scaling cost of ECORD.

\paragraph{Theoretical complexity}
The runtime complexity of any optimisation trajectory scales linearly with the number of actions.  In CO problems where every vertex must be correctly labelled, such as \maxcut{}, episode length scales at best linearly with the number of nodes.
A typical approach, which encapsulates both S2V-DQN and ECO-DQN, for applying RL to CO on graphs parameterises the policy or value function with a GNN.  The precise complexity of a GNN depends on the chosen architecture and graph topology, however, typically the per-layer performance scales linearly with the number of edges in the graph, as this is the number of unique communication channels along which information must be passed.  In practice, this results in a (worst case and typical) polynomial scaling of $\bigO{\abs*{V}^{2}}$ per-action and $\bigO{\abs*{V}^{3}}$ per optimisation, which makes even modest sized graphs with the order of hundreds of nodes very computationally expensive in practice.

In contrast, ECORD only runs the GNN once, regardless of the graph size or episode length, and then selects actions without additional message passing between nodes.  As a result, per-action computational cost scales linearly as $\bigO{\abs*{V}}$, regardless of the graph topology, and the entire optimisation scales as $\bigO{\abs*{V}^{2}}$.
Moreover, action selection in ECORD also has a far smaller memory footprint than using a GNN over the entire graph, and each vertex can be processed in parallel up to the limits of hardware.  Therefore, in practice, we typically obtain a constant, $\bigO{1}$, scaling of the per-action computational cost and, as the single graph network pass is typically negligible compared to the long exploratory phase, complete the entire optimisation in $\bigO{\abs*{V}}$.

\paragraph{Practical performance}
Figure \ref{fig:performance_comparison} demonstrates the practical performance realised from these theoretical improvements.  The left plot compares the action time of ECORD and ECO-DQN on graphs with up to \SI{500}{k} vertices.  Whilst ECORD is always faster, both take near constant time for small graphs where batched inference across all nodes is still efficient.  However, even when ECO-DQN begins to increase in cost ($\abs*{V}{\approx}500$) and eventually fills the GPU memory ($\abs*{V}{\approx}\SI{10}{k}$), ECORD retains a fixed low-cost inference which only appreciably begins to increase for large graphs ($\abs*{V} {>} \SI{100}{k}$).

Moreover, the reduced memory footprint of ECORD also allows for more (randomly initialised) optimisation trajectories to be run in parallel.  The right plot in figure \ref{fig:performance_comparison} contrasts ECORD and ECO-DQN by running as many parallel trajectories as possible for a single graph with up to $\abs*{V}{=}\SI{10}{k}$, and plotting the effective step time (action selection plus environmental step) per trajectory.  The practical increase in throughput of ECORD compared to ECO-DQN increases with graph size from a modest $1.8
{\times}$ with $\abs*{V}{=}30$ to $100{\times}$ for $\abs*{V}{=}\SI{3}{k}$  and $298.9{\times}$ for $\abs*{V}{=}\SI{100}{k}$.

\subsection{Practical performance of ECORD}
\label{sec:gset}

\paragraph{Methods}
ECORD is trained on graphs with binary edge weights, $w(e_{ij}) \in \{0,1\}$ and $\abs{V}=500$.  The optimal parameters are selected according to the performance of the greedy policy on the generated ER10000 set.  For the computational reasons discussed previously, ECO-DQN was trained on ER200 graphs with binary edge weights, and used ER500 graphs for model selection.  The temperature of MCA-soft was tuned independently on each graph, as detailed in \smref{Section D}.

\begin{wrapfigure}[13]{R}{0.45\textwidth}
\vspace{-1.5\baselineskip}
	\centering
	  \includegraphics[width=0.43\textwidth]{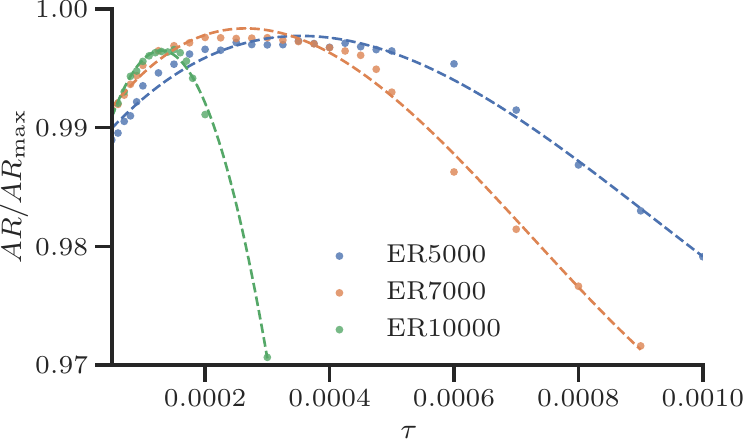}
	  \caption{
	  Effect of temperature on ECORD's performance for $|V|{=}\{\SI{5}{k}, \SI{7}{k}, \SI{10}{k}\}$ graphs.
	  The dashed lines are simply to guide the eye.
	  }
	  \label{fig:temperature_scan}
\end{wrapfigure}

\paragraph{Stochastic exploration}
Despite ECORD using a stochastic soft-greedy policy during training, at test time we previously used a deterministic policy (which, ultimately, still provided near optimal performance on small to medium graphs).
To investigate the performance of a soft-greedy policy, we evaluated ECORD on the large-graph datasets ER5000, ER7000 and ER10000 --  using 20 trajectories of $4\abs{V}$ steps for each graph -- across a range of temperatures.

The results are summarised in \cref{fig:temperature_scan}.  The key points are that a non-zero temperature is optimal for larger graphs, and that the optimal temperature decreases with graph size.  The interpretation is that even an exploratory agent can still eventually get stuck in closed trajectories when acting deterministically.  Therefore, introducing some stochasticity allows ECORD to escape these regions of the solution space and ultimately continue to search for improved configurations.  However, this must be balanced with the need for a sufficiently deterministic policy that a good solution is found with high probability.  Intuitively, longer sequences of actions are required to reach a good configuration for larger graphs, which aligns with the  observed dependence of the optimal temperature on  graph size.

\paragraph{GSet results}
We validate ECORD's performance on large graphs with up to $\num{10000}$ vertices by testing it on all GSet graphs with random binary edges, as summarised in \tabref{tab:gset}.  For each graph we run 20 parallel optimisation trajectories for a maximum of \SI{180}{\second}.  For G55, G60 and G70 we use tuned temperatures, $\tau$, of \num{3.5e-4}, \num{2e-4} and \num{1e-4} respectively, and set $\tau{=}\num{5e-4}$ for all other graphs.

ECORD shows consistently strong performance, matching or beating ECO-DQN and MCA-soft on every instance by a margin that increases with graph size.
The fast exploration of ECORD is emphasised by the fact that it obtains approximation ratios of above \num{0.99} on all graphs with $\abs*{V}\leq\SI{2}{k}$ in under \SI{10}{\second}.
Whilst in principle it is possible to reach the optimal solution in $\abs*{V}$ actions, on larger graphs the obtained cut values consistently increase with longer exploration (for reference, ECORD takes ${>}\SI{28}{k}$ actions in \SI{180}{\second} when $\abs*{V}{=}\SI{10}{k}$). This demonstrates that ECORD learns to search through the solution space without getting stuck, rather than generating a single mode of solutions.

A natural final question is whether the slight decrease in performance on the largest graphs is due to an insufficient time budget, or the learnt policy being less suitable for these problems.  To test this, ECORD was allowed to continue optimising G70 for \SI{1}{\hour}.  Whilst the obtained solution improved to have an approximation ratio of \num{0.978} (obtained in only \SI{233}{\second}), the optimal solution was not found.
Further increasing graph size would be computationally feasible, however G70 is the largest graph in the GSet from the distribution on which we trained.
Ultimately, whilst previous RL-GNN methods were limited by the scalability of GNNs, now we find ourselves limited by the ability of the agent to reason about larger systems -- opening the door to, and defining a challenge for, future research.

\begin{table*}[t]
	\caption{
	Performance on the GSet graphs given a time budget of $10$, $30$, and \SI{180}{\second} (best in \textbf{bold}).
	}
	\vskip 0.15in
	
	\small
	\centering
	\scalebox{\tabscale}{%
	\begin{tabular}{cc|ccc|ccc|ccc}
	
	\toprule
	
	\multicolumn{2}{c}{} &
	\multicolumn{3}{c}{MCA-soft} &
	\multicolumn{3}{c}{ECO-DQN} &
	\multicolumn{3}{c}{ECORD}
	\\
	
	\cmidrule[0.75pt](lr){3-5}
	\cmidrule[0.75pt](lr){6-8}
	\cmidrule[0.75pt](lr){9-11}
	
	Graph & $\abs{V}$ 
	& \SI{10}{\second} & \SI{30}{\second} & \SI{180}{\second} 
	& \SI{10}{\second} & \SI{30}{\second} & \SI{180}{\second}
	& \SI{10}{\second} & \SI{30}{\second} & \SI{180}{\second}
	\\
	
	\cmidrule[0.75pt](lr){1-11}
	
	G1 & \num{800} &
	$0.994$ & $0.999$ & $0.999$ & 
	$0.993$ & $0.993$ & $0.998$ & 
	$0.998$ & $\mathbf{1.000}$ & ---
	\\
	
	G2 & \num{800} &
	$0.995$ & $0.996$ & $0.998$ &
	$0.986$ & $0.995$ & $0.996$ & 
	$0.998$ & $0.998$ & $\mathbf{1.000}$
	\\
	
	G3 & \num{800} &
	$0.996$ & $0.999$ & $0.999$ & 
	$0.992$ & $0.993$ & $0.996$ & 	
	$0.999$ & $\mathbf{1.000}$ & ---
	\\
	
	G4 & \num{800} &
	$0.996$ & $0.997$ & $0.999$ & 
	$0.985$ & $0.992$ & $0.998$ & 
	$0.999$ & $\mathbf{1.000}$ & ---
	\\
	
	G5 & \num{800} &
	$0.996$ & $0.997$ & $0.997$ & 
	$0.993$ & $0.994$ & $0.998$ & 
	$\mathbf{1.000}$ & --- & ---
	\\
	
	\cmidrule[0.25pt](lr){1-11}
	
	G43 & \num{1000} &
	$0.990$ & $0.997$ & $\mathbf{1.000}$ & 
	$0.985$ & $0.992$ & $0.998$ & 
	$0.996$ & $0.997$ & $\mathbf{1.000}$
	\\
	
	G44 & \num{1000} &
	$0.992$ & $0.995$ & $0.999$ & 
	$0.987$ & $0.994$ & $0.996$ & 
	$0.999$ & $0.999$ & $\mathbf{1.000}$
	\\
	
	G45 & \num{1000} &
	$0.995$ & $0.998$ & $0.998$ & 
	$0.988$ & $0.992$ & $0.996$ & 
	$0.998$ & $0.999$ & $\mathbf{0.999}$
	\\
	
	G46 & \num{1000} & 
	$0.992$ & $0.994$ & $0.998$ & 
	$0.990$ & $0.992$ & $0.997$ & 
	$0.998$ & $0.999$ & $\mathbf{1.000}$
	\\
	
	G47 & \num{1000} &
	$0.996$ & $0.997$ & $\mathbf{0.998}$ & 
	$0.988$ & $0.993$ & $0.994$ & 
	$0.997$ & $0.998$ & $\mathbf{0.998}$
	\\
	
	\cmidrule[0.25pt](lr){1-11}
	
	G22 & \num{2000} &
	$0.985$ & $0.990$ & $0.995$ & 
	$0.966$ & $0.981$ & $0.991$ & 
	$0.990$ & $0.995$ & $\mathbf{0.997}$
	\\
	
	G23 & \num{2000} &
	$0.985$ & $0.992$ & $0.996$ & 
	$0.967$ & $0.983$ & $0.989$ & 
	$0.991$ & $0.995$ & $\mathbf{0.997}$
	\\
	
	G24 & \num{2000} &
	$0.986$ & $0.991$ & $\mathbf{0.997}$ &
	$0.967$ & $0.983$ & $0.989$ & 
	$0.992$ & $0.994$ & $\mathbf{0.997}$
	\\
	
	G25 & \num{2000} &
	$0.985$ & $0.993$ & $0.997$ & 
	$0.966$ & $0.985$ & $0.990$ & 
	$0.991$ & $0.996$ & $\mathbf{0.998}$
	\\ 
	 
	G26 & \num{2000} &
	$0.987$ & $0.993$ & $0.995$ &
	$0.963$ & $0.982$ & $0.984$ & 
	$0.991$ & $0.996$ & $\mathbf{0.999}$
	\\
	
	\cmidrule[0.25pt](lr){1-11}
	
	G55 & \num{5000} &
	$0.952$ & $0.968$ & $0.969$ &
	$0.832$ & $0.873$ & $0.939$ & 
	$0.947$ & $0.969$ & $\mathbf{0.985}$
	\\ 
	
	\cmidrule[0.25pt](lr){1-11}
	
	G60 & \num{7000} &
	$0.934$ & $0.963$ & $0.968$ &
	$0.764$ & $0.847$ & $0.930$ & 
	$0.916$ & $0.960$ & $\mathbf{0.981}$
	\\  
	
	\cmidrule[0.25pt](lr){1-11}
	
	G70 & \num{10000} &
	$0.865$ & $0.934$ & $0.957$ &
	$0.677$ & $0.818$ & $0.898$ & 
	$0.837$ & $0.944$ & $\mathbf{0.972}$ 
	\\ 
	 
	\bottomrule
	
	\end{tabular}
	} 
	\label{tab:gset}
\vskip -0.1in
\end{table*}

\paragraph{Ablations}
ECORD's key components are:
(\romannumeral 1)~the use of a single (GNN) encoding step to embed the problem structure,
(\romannumeral 2)~rapid decoding steps where per-vertex actions are conditioned on only local observations and an (RNN) learnt embedding of the optimisation trajectory,
and (\romannumeral 3)~an exploratory CO setting where actions can be reversed.
Ablations (detailed in  \smref{Section E}) of the GNN and RNN show that both are necessary for strong performance.
Interestingly, whilst removing the GNN has limited impact on small instances, on 2/3 of the larger ($\abs{V}{>}\SI{2}{k}$) instances it is, in fact, the most impactful ablation. As a likely utility of the GNN is to encode priors on the absolute or relative labels of each vertex in the solution, it seems reasonable that this becomes more important to guide the search on larger instances (where a smaller fraction of the search space can be explored within a fixed time budget).
ECORD's contribution to the exploratory CO setting is evidencing that a suitably stochastic policy outperforms a deterministic one.  To emphasise that this is not simply because of algorithmic improvements in M-DQN compared to DQN, a deterministic-acting ECORD was trained with DQN and found to not match ECORD with M-DQN (see \smref{Section E}).

\section{Discussion}

We present ECORD, a new \sota{} RL algorithm for \maxcut{} in terms of both performance and scalability.  ECORD's demonstrated efficacy on graphs with up \SI{10}{k} vertices, and highly favourable computational complexity, suggests even larger problems could be tackled.  By replacing multiple expensive GNN operations with a single embedding stage and rapid action-selection directed by a recurrent unit, this work highlights the importance of, and a method to achieve, efficient exploration when solving CO problems, all within the broader pursuit of scalable geometric~learning.

To the best of our knowledge the graphs scaled to in this work are the largest successfully considered by learnt heuristics for \maxcut{} -- with ECO-DQN~\citep{barrett20}, S2V-DQN~\citep{khalil17} and the RL-SA algorithm of \citet{beloborodov2020reinforcement} previously scaling to \SI{2}{k}, \SI{1.2}{k} and \num{800} vertices, respectively.
Full comparison to the numerous non-ML based approaches to CO problems is beyond the scope of this paper (though extended baselines are presented in \smref{Section E}), but it is clear that ECORD represents a significant step forward in the scalability of learnt heuristics.

Algorithmic improvements are a possible direction for further research.  
An adaptive (or learnt) temperature schedule could better trade-off stochastic exploration and deterministic solution improvement.
As ECORD runs multiple optimisations in parallel, utilising information from other trajectories to either better inform future decisions, or ensure sufficient diversity between them, would be another approach for improving performance.
An alternative direction would be to tackle other problems and, in principle, ECORD could be applied to any vertex-labelling (graph partitioning) CO problem defined on a graph.  However, its use of `peeks' (one-step look aheads) does not directly translate to problems where the quality of intermediate actions are not naturally evaluated (\eg{} those where certain node labelings may be invalid such as maximum clique).  Considering only valid actions or utilising indirect metrics of solution quality are possible remedies, but this remains a topic for future~research.
%
%

\bibliography{references}
\bibliographystyle{plainnat}

\newpage

\appendix
\section{Training details}

The ECORD pseudocode is given in algorithm \ref{alg:ecord}. Note that to update the network at time $t$ we make use of truncated backpropagation through time (BPTT) \citep{werbos1990backpropagation, sutskever2013training} using the previous $t_{\text{BPTT}}$ experiences. We perform network updates every $f_{\text{upd}}$ time steps, where we update the online network parameters $\theta$ using stochastic gradient descent (SGD) on the M-DQN loss and the target network parameters $\overline{\theta}$ using a soft update \citep{lillicrap2015continuous} with update parameter $\tau_{\text{upd}}$. The hyperparameters used are summarised in Table \ref{tab:parameters}.

\begin{algorithm}
	\caption{Training ECORD} 
	\label{alg:ecord}
\begin{algorithmic}
	\STATE Initialise experience replay memory $\mathcal{M}$.
	
	\FOR{each batch of episodes}
		
	\STATE Sample $B\sub{G}$ graphs $G(V,E)$ from distribution $\mathbb{D}$\;
	\STATE Calculate per-vertex embeddings using GNN\;
	\STATE Initialise a random solution set for each graph, $S_0 \subset V$\;
	
	\FOR{each step $t$ in the episode}
	
	\STATE $k^{\prime} = \min(t-k\sub{BPTT}, 0)$\;
	
	\FOR{each graph $G_j$ in the batch}
	
	\STATE $a^{(t)} {\sim} 
		\begin{cases}
 			\text{randomly from } V \text{ with prob.\ }\varepsilon \\
    		\softmax(\tfrac{Q_\theta(s^{(t)},\cdot)}{\tau})           \text{ with prob.\ }1-\varepsilon
    	\end{cases}$\;%
 	\STATE $S^{(t+1)} :=
		\begin{cases}
   			S^{(t)} \cup \{a^{(t)}\},& \text{if }a^{(t)} \notin S^{(t)}  \\
    		S^{(t)}  \setminus \{a^{(t)}\},& \text{if }a^{(t)} \in S^{(t)}  \\
		\end{cases}$\;
	\STATE Add tuple
		$m^{(t)} = (
		s^{(t-k^{\prime})},\dots,s^{(t+1)},
		a^{(t-k^{\prime})},\dots,a^{(t)},
		r^{(t)},
		d
		)$ to $\mathcal{M}$\;
	\IF{$t\bmod f\sub{upd}==0$}
	\STATE Sample batch of experiences from buffer, $M^{(t)} \subset \mathcal{M}$
	\STATE Update $\theta$ with one SGD step using BPTT from $t$ down to $t-k^{\prime}$ on $\mathcal{L}\sub{m-dqn}$
	\STATE Update target network, $\overline{\theta} \leftarrow \theta\tau_{\text{upd}} + \overline{\theta}(1-\tau_{\text{upd}})$
	\ENDIF
	
	\ENDFOR 
	
	\ENDFOR 
	
	\ENDFOR 
%
\end{algorithmic}
\end{algorithm}

\begin{table*}[h]
\caption{Parameters used for ECORD unless otherwise stated.}
\vskip 0.15in

\small
\centering
\begin{tabular}{lr}

\toprule

Parameter & Value \\

\cmidrule[0.75pt]{0-1}

Number of training steps & \num{40000} \\
Batch size & 64 \\
Update frequency ($f\sub{upd}$) & 8 \\
Learning rate & $\num{1e-3}$ \\
Optimizer & Adam ($\beta_1=0.9$, $\beta_2=0.999$) \\
Soft update rate ($\tau\sub{upd}$) & 0.01 \\

BPTT length ($k\sub{BPTT}$) & 5 \\
Buffer size ($\abs{\mathcal{M}}$) & \num{40000} \\
Discount factor ($\gamma$) & $0.7$ \\

Initial exploration probability ($\varepsilon^{(t=0)}$) & 1 \\
Final exploration probability ($\varepsilon^{(t_{\varepsilon})}$) & 0.05 \\
Time of exploration decay ($t_{\varepsilon}$) & \num{5000} \\

M-DQN temperature ($\tau$) & 0.01 \\
M-DQN bootstrap ($\alpha$) & 0.9 \\
M-DQN clipping ($l_0$) & -1 \\

\bottomrule
	
\end{tabular}
\label{tab:parameters}
\vskip -0.1in
\end{table*}

\section{Architecture details}
\label{app:architecture}

Here we provide further details on the network architecture described in Section 3.2 and Figure 1 of the main text.  For simplicity, we implicitly drop any bias terms in the below equations.

\paragraph{Graph Neural Network}
The gated graph convolution network of \citet{li15} is modified to include layer normalisation, $\layernorm(\cdot)$, where the 16-dimensional embedding of node $i$ at layer $l+1$ is given by,
\begin{equation}
	x_i^{(l+1)} = \layernorm(\mathrm{GRU}(x_i^{(l)}, m_i^{(l+1)})), \quad
	m_i^{(l+1)} = \frac{1}{\abs{N(i)}}\sum_{j \in \mathcal{N}(i)} w(e_{ij}) W\sub{g} x_j^{(l)},
\end{equation}
where $	W\sub{g} \in \Rab{16}{16}$.
The final embeddings after 4 rounds of message-passing are linearly projected to per-vertex embeddings, $x_i = W\sub{p} x_i^{(4)}$ with $W\sub{p} \in \Rab{16}{16}$, that remain fixed for the remainder of the episode.

\paragraph{Value network}
 Recall from the main text that Q-values at time $t$ are predicted as
 \begin{equation}
	Q(v_i^{(t)}, h^{(t)})  = V(h^{(t)}) + A(v_i^{(t)}, h^{(t)}) = \text{MLP}\sub{V}(h^{(t)}) + \text{MLP}\sub{A}([v_i^{(t)}, W\sub{h} h^{(t)}]).
\end{equation}
where $h^{(t)} \in \R^{1024}$ is the hidden state of the RNN and $v_i^{(t)} = [ x_i, W\sub{o} o_i^{(t)} ] \in \R^{32}$ are the per-node embeddings with $x_i \in \R^{16}$, $ W\sub{o} \in \Rab{16}{3}$ and $W\sub{h} \in \Rab{32}{1014}$.

The value head, $\text{MLP}\sub{V}(\cdot): \R^{1024} \rightarrow \R$, is a 2-layer network that applies a $\tanh$ activation to the input and leaky ReLU to the intermediate activations.

The advantage head, $\text{MLP}\sub{A}(\cdot): \R^{64} \rightarrow \R$, is a 2-layer network that applies layer norm and leaky ReLU to the intermediate activations.

\paragraph{Recurrent network}
The RNN hidden state is 1024-dimensional and initialised to zeros.  It is updated using the embedding of the selected action, $v_{\ast}^{(t)}$, and our global observation from the next step,
\begin{equation}
	h^{(t+1)} = \text{GRU}(h^{(t)}, m^{(t+1)}), \quad \text{where} \quad
	m^{(t+1)} = \text{LeakyReLU}(W\sub{m}([v_{\ast}^{(t)}, o\sub{G}^{(t+1)}])),
\end{equation}
where $W\sub{m} \in \Rab{64}{34}$.

\section{Efficient re-calculation of cut value and peeks}
\label{sec:app_peeks}

Given a graph, $G(V,E)$, with edge weights $w_{ij} \equiv w(e_{ij})$ for $e_{ij} \in E$, and a node labelling represented as a binary vector, $z \in \{0,1\}^{\abs*{V}}$, the cut value is given by,
\begin{equation}
\begin{aligned}
	C(z \vert G) &= \frac{1}{2} \sum_{ij} w_{ij} \left( z_i(1-z_j) + (1-z_i)z_j \right), \\
	&= \frac{1}{2} \sum_{ij} w_{ij} (z_i + z_j - 2 z_i z_j).
\end{aligned}
\label{eq:cutVal}
\end{equation}
It is straightforward to decompose this into the sum of `local' cuts
\begin{equation}
	C(z \vert G) = \frac{1}{2} \sum_i C_i,
	\quad
	C_i = \sum_{j \in \mathcal{N}(i)} w_{ij} (z_i + z_j - 2 z_i z_j),
\end{equation}
where $C_i$, the cut value of the sub-graph containing node only $i$ and its neighbours, $\mathcal{N}(i)$.  Similarly, we can define the total weight of un-cut edges connected to each vertex as
\begin{equation}
	\overline{C}_i = \sum_{j \in \mathcal{N}(i)} w_{ij} (1 - z_i - z_j + 2 z_i z_j).
\end{equation}
The change in cut value (referred to as the `peek' feature in the main text) if the label of vertex $i$ is flipped is then then given by
\begin{equation}
\begin{aligned}
	\Delta C_i &= \overline{C}_i - C_i, \\
	&= \sum_{j \in \mathcal{N}(i)} w_{ij} (4 z_i z_j - 2(z_i + z_j) + 1), \\
	&= \sum_{j \in \mathcal{N}(i)} w_{ij} (2 z_i - 1) (2 z_j - 1).
\end{aligned}
\end{equation}
Calculating these one-step look-aheads to the change in cut for each action clearly has the same complexity as calculating the cut-value itself (equation (\ref{eq:cutVal})).  Moreover, they only have to be calculated once at the start of each episode, as when vertex $i$ is flipped from $z_i$ to $\overline{z}_i$, only the `peeks' of vertrices $i$ and $j \in \mathcal{N}(i)$ need to be updated.  These updates follow directly from the above and are given by
\begin{equation}
	\Delta C_i \rightarrow - \Delta C_i, \quad \Delta C_j \rightarrow \Delta C_j - w_{ij} (2 z_i - 1) (2 z_j - 1).
	\label{eq:peeks}
\end{equation}

\section{MCA-soft}
\label{sec:mca-soft}

MCA-soft attempts to upper bound the performance simple policies that condition actions based solely on the provided `peeks' for each action.  Denoting the known change in cut value if vertex $i$ was to be flipped as $\Delta C_i$ (see eq.~(\ref{eq:peeks})), MCA-soft follows a stochastic policy given by
\begin{equation}
	a^{(t)} \sim \softmax\left(\frac{ \Delta C_i }{ \tau\sub{mca} }\right).
\end{equation}
To maximise the performance of MCA-soft, the temperature, $\tau\sub{mca} \in \R$, is independently tuned to maximise performance on every set of graphs considered.  In practice, this process is a grid search over $\tau\sub{mca} \in \{ 0, 0.001, 0.003, 0.01, 0.03, 0.1, 0.3, 1, 3 \}$ for results in Table 1 (main text) and \ref{tab:sota_comp_2} (SM), and $\tau\sub{mca} \in \{ 0, 0.0001, 0.001, 0.01, 0.1, 1\}$ for results in Table 2 (main text).

\section{Extended results}
\label{sec:extened_results}

\paragraph{Training on BA graphs}
For completeness, we repeat the experiments shown in the main manuscript but now training on $40$-vertex BA graphs and evaluating on both ER and BA graphs with up to $500$ vertices, with the results shown in Table \ref{tab:sota_comp_2}.

\begin{table*}[h]
\centering
\caption{
Scores for agents trained on BA40.  Error bars denote \SI{68}{\percent} confidence intervals. RL baseline scores are taken directly from~\citet{barrett20} to provide the fairest possible comparison. ECORD results averaged across 3 seeds.
}
\vskip 0.15in

\scalebox{\tabscale}{%
\begin{tabular}{c cc cc c}
\toprule

\multicolumn{1}{c}{} &
\multicolumn{2}{c}{Heuristics} &
\multicolumn{2}{c}{RL baselines} &
\multicolumn{1}{c}{}
\\

\cmidrule[0.75pt](lr){2-3}
\cmidrule[0.75pt](lr){4-5}

& MCA & MCA-soft & S2V-DQN & ECO-DQN & ECORD \\

\cmidrule[0.75pt](lr){1-6}

BA40 &
 $0.999\pmstack{0.000}{0.006}$ & 
 $\mathbf{ 1.000 }\pmstack{0.000}{0.000}$ & 
 $0.961\pmstack{0.027}{0.048}$ &
 $\mathbf{ 1.000 }\pmstack{0.000}{0.000}$ & 
 $\mathbf{ 1.000 }\pmstack{0.000}{0.000}$ \\
BA60 &
 $0.989\pmstack{0.007}{0.018}$ & 
 $0.997\pmstack{0.008}{0.008}$ & 
 $0.959\pmstack{0.030}{0.040}$ &
 $\mathbf{ 1.000 }\pmstack{0.000}{0.000}$ &
 $\mathbf{ 1.000 }\pmstack{0.000}{0.000}$ \\
BA100 &
 $0.965\pmstack{0.020}{0.020}$ &
 $0.984\pmstack{0.012}{0.013}$ &
 $0.941\pmstack{0.037}{0.044}$ &
 $\mathbf{ 1.000 }\pmstack{0.000}{0.001}$ &
 $\mathbf{ 1.000 }\pmstack{0.000}{0.001}$ \\
BA200 &
 $0.911\pmstack{0.033}{0.037}$ &
 $0.929\pmstack{0.034}{0.034}$ &
 $0.808\pmstack{0.107}{0.102}$ &
 $\mathbf{ 0.983 }\pmstack{0.009}{0.034}$ &
 $0.980\pmstack{0.020}{0.033}$ \\
BA500 &
 $0.889\pmstack{0.015}{0.015}$ &
 $0.899\pmstack{0.014}{0.014}$ & 
 $0.499\pmstack{0.114}{0.114}$ &
 $\mathbf{ 0.990 }\pmstack{0.008}{0.008}$ &
 $0.967\pmstack{0.012}{0.012}$ \\

\cmidrule[0.5pt](lr){1-6}

ER40 &
 $0.997\pmstack{0.001}{0.010}$ & 
 $\mathbf{ 1.000 }\pmstack{0.000}{0.000}$ & 
 $0.970\pmstack{0.020}{0.037}$ &
 $\mathbf{ 1.000 }\pmstack{0.000}{0.000}$ & 
 $\mathbf{ 1.000 }\pmstack{0.000}{0.000}$ \\
ER60 &
 $0.994\pmstack{0.003}{0.013}$ & 
 $0.999\pmstack{0.001}{0.004}$ & 
 $0.951\pmstack{0.036}{0.041}$ &
 $\mathbf{ 1.000 }\pmstack{0.000}{0.000}$ &
 $\mathbf{ 1.000 }\pmstack{0.000}{0.000}$ \\
ER100 &
 $0.977\pmstack{0.017}{0.017}$ &
 $0.993\pmstack{0.007}{0.008}$ &
 $0.941\pmstack{0.035}{0.037}$ &
 $\mathbf{ 1.000 }\pmstack{0.000}{0.001}$ &
 $\mathbf{ 1.000 }\pmstack{0.000}{0.001}$ \\
ER200 &
 $0.959\pmstack{0.014}{0.014}$ &
 $0.973\pmstack{0.012}{0.012}$ &
 $0.933\pmstack{0.024}{0.024}$ &
 $0.994\pmstack{0.004}{0.005}$ &
 $\mathbf{ 1.000 }\pmstack{0.000}{0.001}$\\
ER500 &
 $0.941\pmstack{0.012}{0.012}$ &
 $0.958\pmstack{0.009}{0.009}$ & 
 $0.905\pmstack{0.019}{0.019}$ &
 $0.979\pmstack{0.006}{0.006}$ &
 $\mathbf{ 0.996 }\pmstack{0.004}{0.004}$ \\
  
\bottomrule

\end{tabular}
} 
\label{tab:sota_comp_2}
\vskip -0.1in
\end{table*}

\paragraph{Baseline comparison}
Here we conduct a thorough solver comparison by training and inferring on both ER and BA graphs of up to $500$ vertices. We compare ECORD to seven baselines; six taken directly from \citet{barrett20}\footnote{Code and data available in paper or at \url{https://github.com/tomdbar/eco-dqn} under an MIT license.} (ECO-DQN, S2V-DQN, MCA, CPLEX, SimCIM, and \citet{leleu19}; refer to \citet{barrett20} for implementation details), and one an extension of MCA (MCA-soft, as described in this manuscript). The results are summarised in Table \ref{tab:full_sota_comp}, with the approximation ratios shown taken from averaging the solvers' performances across $100$ BA and ER graphs of up to $500$ vertices.

\begin{table*}[h]
\centering
\caption{
Comparison of the approximation ratios for the Greedy (MCA and MCA-soft), branch-and-bound (CPLEX), simulated annealing (SimCIM and Leleu), and RL (S2V-DQN and ECO-DQN) \maxcut{} solver baselines.
}
\vskip 0.15in
\scalebox{\tabscale}{%
\begin{tabular}{c c c c c c c c c}

\toprule

& MCA & MCA-soft & CPLEX & SimCIM & Leleu & S2V-DQN & ECO-DQN & ECORD \\

\hline

ER40 &
 $0.997$& 
 $1.000$ & 
 $1.000$ &
 $1.000$ & 
 $1.000$ &
 $0.980$ &
 $1.000$ &
 $1.000$ \\
ER60 &
 $0.994$ & 
 $0.999$ & 
 $1.000$ &
 $1.000$ &
 $1.000$ &
 $0.973$ &
 $1.000$ &
 $1.000$ \\
ER100 &
 $0.977$ &
 $0.993$ &
 $0.870$ &
 $1.000$ &
 $1.000$ &
 $0.961$ &
 $1.000$ &
 $1.000$ \\
ER200 &
 $0.959$ &
 $0.973$ &
 $0.460$ &
 $0.990$ &
 $1.000$ &
 $0.951$ &
 $0.999$ &
 $1.000$ \\
ER500 &
 $0.941$ &
 $0.958$ & 
 $0.160$ &
 $0.990$ &
 $1.000$ &
 $0.921$ &
 $0.985$ &
 $0.996$ \\
 
 \hline
 
BA40 &
 $0.999$ & 
 $1.000$ & 
 $1.000$ &
 $1.000$ & 
 $1.000$ &
 $0.961$ &
 $1.000$ &
 $1.000$ \\
BA60 &
 $0.989$ & 
 $0.997$ & 
 $1.000$ &
 $1.000$ &
 $1.000$ &
 $0.959$ &
 $1.000$ &
 $1.000$ \\
BA100 &
 $0.965$ &
 $0.984$ &
 $1.000$ &
 $0.990$ &
 $1.000$ &
 $0.961$ &
 $1.000$ &
 $0.996$ \\
BA200 &
 $0.911$ &
 $0.929$ &
 $0.830$ &
 $0.990$ &
 $0.940$ &
 $0.808$ &
 $0.983$ &
 $0.980$ \\
BA500 &
 $0.889$ &
 $0.899$ & 
 $0.170$ &
 $0.970$ &
 $1.000$ &
 $0.499$ &
 $0.990$ &
 $0.967$ \\
  
\bottomrule

\end{tabular}
} 
\label{tab:full_sota_comp}
\vskip -0.1in
\end{table*}

\paragraph{Semidefinite programming comparison}
In addition to heuristics, another important branch of \maxcut{} solver research is that of approximation algorithms. Such algorithms can offer a theoretical guarantee on the approximation ratio while still solving a relaxed formulation of the original problem in polynomial time. One such approximation method is the canonical semidefinite programming (SDP) approach of \citet{goemans95}. \citet{goemans95} first formulate \maxcut{} as an SDP by framing the objective as a linear function of a symmetric matrix subject to linear equality constraints (as in a linear programme) but with the additional constraint that the matrix must be positive semidefinite (whereby, for an $n \times n$ matrix $A$, $\forall x \in \mathcal{R}^{n}, x^{t} Ax \geq 0$). This relaxed \maxcut{} SDP formulation can be solved efficiently using algorithms such as a generalised Simplex method. The insight of \citet{goemans95} was to then apply a geometric \textit{randomised rounding} technique to convert the SDP solution into a feasible \maxcut{} solution. Crucially, the randomised rounding method gives a guarantee to be within at least $0.87856$ times the optimal \maxcut{} value. 

To the best of our knowledge, no open-access \citet{goemans95} solver exists which can handle \maxcut{} problems with negatively weighted edges. However, G1-5 of the GSet graphs used in Table 2 (main text) of this manuscript have all-positive edge weights, therefore we ran the open-source cvx solver (\url{https://github.com/hermish/cvx-graph-algorithms}), which implements \citet{goemans95}, on these $5$ problems to obtain approximation ratios of $0.971, 0.970, 0.975, 0.970,$ and $0.967$ in $449, 484, 496, 521,$ and \SI{599}{\second} respectively. In addition to ECORD outperforming \citet{goemans95} on these $5$ GSet graphs (see Table 2 (main text)) in both solving time and optimality, we note that ECORD also exceeds the $0.87856$ approximation ratio guarantee across all the GSet, ER, and BA graphs examined in our work (see Tables 1 and 2 (main text) and \ref{tab:sota_comp_2}, and \ref{tab:full_sota_comp} (SM)).

\paragraph{Ablations}
We provide ablations to further investigate the key components of ECORD, highlighted in the main text as
\begin{enumerate}
    \item a GNN to encode the spatial structure of the problem,
    \item a rapid decoding conditioned on local per-node observations and an RNN's internal state that represents the optimisation trajectory,
    \item an exploratory CO setting.
\end{enumerate}
We ablate the GNN and RNN by using fixed zero-vectors in-place of the static per-node embeddings, $x_i$, and the RNN hidden state, $h^{(t)}$, respectively.  As the advantage of exploratory CO has already been demonstrated in prior works~\citep{barrett20}, we instead ablate the stochastic policies we find provide further improved exploration at test time (see Section 4.3 of the main text).  Specifically, we train the agent using DQN instead of M-DQN, as this optimises for a deterministic policy.

For all ablations, we otherwise use the same procedure as the full ECORD agent from Section 4.3 of the main text, including tuning temperature of the agents policy with a grid-search.
DQN is, unsurprisingly, found to be best at $\tau=0$, as is the agent with the GNN ablated. 
The agent with the RNN ablated uses $\tau$ of \num{1e-4}, \num{8.5e-5} and \num{5e-5} for For G55, G60 and G70, respectively, with $\tau{=}\num{5e-4}$ for all other graphs.  Results are presented in Table~\ref{tab:gset_ablations} with ECORD significantly outperforming all ablations on larger graphs.

\begin{table*}[!t]
	\caption{
	ECORD, including both a GNN and RNN trained with M-DQN, and ablations (as described in the text) on the GSet graphs given a time budget of $10$, $30$, and \SI{180}{\second} (best in \textbf{bold}).
	}
	\vskip 0.15in

	\small
	\centering
	\resizebox{\textwidth}{!}{%
	\begin{tabular}{cc|ccc|ccc|ccc|ccc}
	
	\toprule
	
	\multicolumn{2}{c}{} &
	\multicolumn{3}{c}{No GNN} &
	\multicolumn{3}{c}{No RNN} &
	\multicolumn{3}{c}{DQN} &
	\multicolumn{3}{c}{ECORD}
	\\
	
	\cmidrule[0.75pt](lr){3-5}
	\cmidrule[0.75pt](lr){6-8}
	\cmidrule[0.75pt](lr){9-11}
	\cmidrule[0.75pt](lr){12-14}
	
	Graph & $\abs{V}$ 
	& \SI{10}{\second} & \SI{30}{\second} & \SI{180}{\second}
	& \SI{10}{\second} & \SI{30}{\second} & \SI{180}{\second} 
	& \SI{10}{\second} & \SI{30}{\second} & \SI{180}{\second}
	& \SI{10}{\second} & \SI{30}{\second} & \SI{180}{\second}
	\\
	
	\cmidrule[0.75pt](lr){1-14}
	
	G1 & \num{800} &
	$\mathbf{1.000}$ & --- & --- & 
	$0.998$ & $\mathbf{1.000}$ & --- & 
	$0.996$ & $0.999$ & $0.999$ & 
	$0.998$ & $\mathbf{1.000}$ & ---
	\\
	
	G2 & \num{800} &
	$0.999$ & $\mathbf{1.000}$ & --- & 
	$0.998$ & $0.998$ & $0.999$ & 
	$0.998$ & $0.998$ & $0.999$ & 
	$0.998$ & $0.998$ & $\mathbf{1.000}$ 
	\\
	
	G3 & \num{800} &
	$0.999$ & $\mathbf{1.000}$ & --- & 
	$0.998$ & $0.999$ & $0.999$ & 
	$0.996$ & $0.999$ & $0.999$ & 
	$0.999$ & $\mathbf{1.000}$ & ---
	\\
	
	G4 & \num{800} &
	$\mathbf{1.000}$ & --- & --- & 
	$0.999$ & $\mathbf{1.000}$ & --- & 
	$0.998$ & $0.999$ & $0.999$ & 
	$0.999$ & $\mathbf{1.000}$ & ---
	\\
	
	G5 & \num{800} &
	$\mathbf{1.000}$ & --- & --- & 
	$0.999$ & $0.999$ & $0.999$ & 
	$0.998$ & $0.998$ & $0.998$ & 
	$\mathbf{1.000}$ & --- & ---
	\\
	
	\cmidrule[0.25pt](lr){1-14}
	
	G43 & \num{1000} &
	$0.997$ & $0.998$ & $0.999$ & 
	$0.996$ & $0.999$ & $0.999$ & 
	$0.991$ & $0.991$ & $0.991$ & 
	$0.996$ & $0.997$ & $\mathbf{1.000}$
	\\
	
	G44 & \num{1000} &
	$0.997$ & $0.998$ & $\mathbf{1.000}$ & 
	$0.999$ & $0.999$ & $0.999$ & 
	$0.995$ & $0.995$ & $0.995$ & 
	$0.999$ & $0.999$ & $\mathbf{1.000}$
	\\
	
	G45 & \num{1000} &
	$0.995$ & $0.997$ & $0.999$ & 
	$0.997$ & $\mathbf{1.000}$ & --- & 
	$0.994$ & $0.994$ & $0.994$ & 
	$0.998$ & $0.999$ & $\mathbf{0.999}$
	\\
	
	G46 & \num{1000} & 
	$0.997$ & $0.998$ & $0.999$ & 
	$0.995$ & $0.996$ & $0.998$ & 
	$0.995$ & $0.995$ & $0.995$ & 
	$0.998$ & $0.999$ & $\mathbf{1.000}$
	\\
	
	G47 & \num{1000} &
	$0.996$ & $0.998$ & $\mathbf{1.000}$ & 
	$0.996$ & $0.996$ & $\mathbf{1.000}$ &
	$0.992$ & $0.993$ & $0.993$ & 
	$0.997$ & $0.998$ & $0.998$
	\\
	
	\cmidrule[0.25pt](lr){1-14}
	
	G22 & \num{2000} &
	$0.990$ & $0.991$ & $0.994$ & 
	$0.990$ & $0.995$ & $0.999$ & 
	$0.987$ & $0.989$ & $0.989$ & 
	$0.990$ & $0.995$ & $\mathbf{0.997}$
	\\
	
	G23 & \num{2000} &
	$0.990$ & $0.995$ & $0.997$ & 
	$0.989$ & $0.993$ & $0.998$ & 
	$0.987$ & $0.991$ & $0.991$ & 
	$0.991$ & $0.995$ & $\mathbf{0.997}$
	\\
	
	G24 & \num{2000} &
	$0.989$ & $0.993$ & $0.998$ & 
	$0.990$ & $0.993$ & $0.996$ & 
	$0.987$ & $0.989$ & $0.989$ & 
	$0.992$ & $0.994$ & $\mathbf{0.997}$
	\\
	
	G25 & \num{2000} &
	$0.991$ & $0.995$ & $0.996$ & 
	$0.989$ & $0.994$ & $0.997$ & 
	$0.988$ & $0.993$ & $0.993$ & 
	$0.991$ & $0.996$ & $\mathbf{0.998}$
	\\ 
	 
	G26 & \num{2000} &
	$0.989$ & $0.994$ & $0.997$ & 
	$0.991$ & $0.994$ & $0.998$ & 
	$0.991$ & $0.994$ & $0.994$ & 
	$0.991$ & $0.996$ & $\mathbf{0.999}$
	\\
	
	\cmidrule[0.25pt](lr){1-14}
	
	G55 & \num{5000} &
	$0.924$ & $0.949$ & $0.951$ & 
	$0.950$ & $0.971$ & $0.981$ & 
	$0.951$ & $0.953$ & $0.956$ & 
	$0.947$ & $0.969$ & $\mathbf{0.985}$
	\\
	
	\cmidrule[0.25pt](lr){1-14}
	
	G60 & \num{7000} &
	$0.796$ & $0.933$ & $0.950$ & 
	$0.900$ & $0.953$ & $0.972$ & 
	$0.918$ & $0.949$ & $0.951$ & 
	$0.916$ & $0.960$ & $\mathbf{0.981}$
	\\
	
	\cmidrule[0.25pt](lr){1-14}
	
	G70 & \num{10000} &
	$0.703$ & $0.703$ & $0.931$ & 
	$0.766$ & $0.837$ & $0.879$ & 
	$0.761$ & $0.929$ & $0.933$ & 
	$0.837$ & $0.944$ & $\mathbf{0.972}$
	\\
	 
	\bottomrule
	
	\end{tabular}
	} 
	\label{tab:gset_ablations}
\vskip -0.1in
\end{table*}

\paragraph{Graph network timing}
As stated in the main text, the single pass of the graph neural network is negligible compared to the overall run time of ECORD.   To be concrete, on the largest graph for which results are reported (G70 with $\abs*{V}=\SI{10}{k}$ nodes), our embedding stage takes \SI{1.96\pm0.09}{\ms}, compared to the tens of seconds of exploratory decoding.

\end{document}